\documentclass{article} 
\usepackage{iclr2016_conference,times}
\usepackage{hyperref}
\usepackage{url}
\usepackage{soul}
\usepackage{dirtytalk}
\usepackage{epigraph}
\usepackage{graphicx}
\usepackage{caption}
\usepackage[noend]{algpseudocode}
\usepackage{subcaption}
\usepackage{censor}
\usepackage{algorithm}
\usepackage{amsmath,esint}

\newcommand*\samethanks[1][\value{footnote}]{\footnotemark[#1]}

\title{Stopping GAN Violence: \\
       Generative Unadversarial Networks}

\author{Samuel Albanie\thanks{Authors are listed according to the degree to 
        which their  home nation underperformed at the 2016 European football 
        championships} \\
Institute of Deep Statistical Harmony\\
Shelfanger, UK\\
\AND
S\'ebastien Ehrhardt\samethanks\\
French Foreign Legion\\
Location Redacted\\
\AND
Jo\~{a}o F. Henriques\samethanks\\
Centre for Discrete Peace, Love and Understanding\\
Coimbra, Portugal\\
}

\begin{document}

\maketitle


\begin{abstract}
While the costs of human violence have attracted a great deal of attention
from the research community, the effects of the network-on-network (NoN) 
violence popularised by Generative Adversarial Networks have yet to be 
addressed. In this work, we quantify the financial, social, 
spiritual, cultural, grammatical and dermatological impact of this aggression 
and address the issue by proposing a more peaceful approach which we term 
\textit{Generative Unadversarial Networks} (GUNs). Under this framework, 
we simultaneously train two models: 
a generator $G$ that does its best to capture whichever data distribution it 
feels it can manage, and a motivator $M$ that helps $G$ to achieve its dream. 
Fighting is strictly \textit{verboten} and both models evolve by learning 
to respect their differences. The framework is both theoretically and 
electrically grounded in game theory, and can be viewed as a 
\textit{winner-shares-all} two-player game in which both players work as a 
team to achieve the best score. Experiments show that by working in harmony, 
the proposed model is able to claim both the moral and log-likelihood high 
ground. Our work builds on a rich history of carefully argued position-papers,
published as anonymous YouTube comments, which prove that the optimal solution 
to NoN violence is more GUNs.
\end{abstract}

%

\epigraph{Takes skill to be real, time to heal each 
other}{\textit{Tupac Shakur, Changes, 1998}}

\section{Introduction}

Deep generative modelling is probably important (see e.g. 
\cite{bengio2013representation}, 
\cite{bengio2013generalized}, 
\cite{bengio2007greedy}, 
\cite{bengio2015deep} 
\cite{bengio2007scaling}
and (Schmidhuber et al., circa 3114 BC)).  
Justifications recently overheard in the nightclubs of Cowley\footnote{The 
nightclubs of Cowley are renowned for their longstanding philosophical 
support for Dubstep, Grime and Connectionism, and should not be confused 
with the central Oxford nightclub collective which leans more towards 
Dubstep, Grime and Computationalism - speak to \textit{Old Man Bridge} at 
3am on a Friday morning under the stairs of the smoking area for a more 
nuanced clarification of the metaphysical differences of opinion.} include
the ability to accurately approximate data distributions without prohibitively
expensive label acquisition, and computationally 
feasible approaches to beating human infants at chess\footnote{Infants of other 
species (fox cubs, for example) remain an adorable open question in the field.}. 
Deep generative modelling was broadly considered intractorable, until recent 
groundbreaking research by \cite{goodfellow2014generative} employed 
machiavellian adversarial tactics to demonstrate that methaphorical tractors 
could in fact be driven directly through the goddamn centre of this previously 
unploughed research field (subject to EU agricultural safety and set-aside 
regulations).  

The key insight behind Generative Adversarial Networks (commonly referred to as 
GANs, GANGs or CAPONEs depending on sources of counterfeit currency) is to 
pit one model against another in a gladiatorial 
quest for dominance. However, as ably illustrated by respected human actor and
philanthropist Russell Crowe in the documentary \textit{Gladiator}, being an 
\textit{actual gladiator} isn't all sunshine and rainbows---although it's 
possible to get a great tan, one still has to wear sandals.

Even though we are only in the introduction, we now bravely leap into a 
series of back-of-the-envelope calculations to compute a lower bound on 
the cost of that violence for the case of middle aged, median-income 
Generative Adversarial Networks living in comfortable, but affordable 
accommodation in the leafy suburbs of an appropriate class of functions. 

Following the literature, we define the adversaries as two models, a 
discriminator $D$ and a generator $G$. However, since we don't agree with 
the literature or wish to condone its violent actions in any form, we 
immediately redefine the models as follows:

\begin{equation}
 D, G := G, D
\end{equation}

 \begin{figure}
    \centering
    \includegraphics[width=0.4\textwidth]{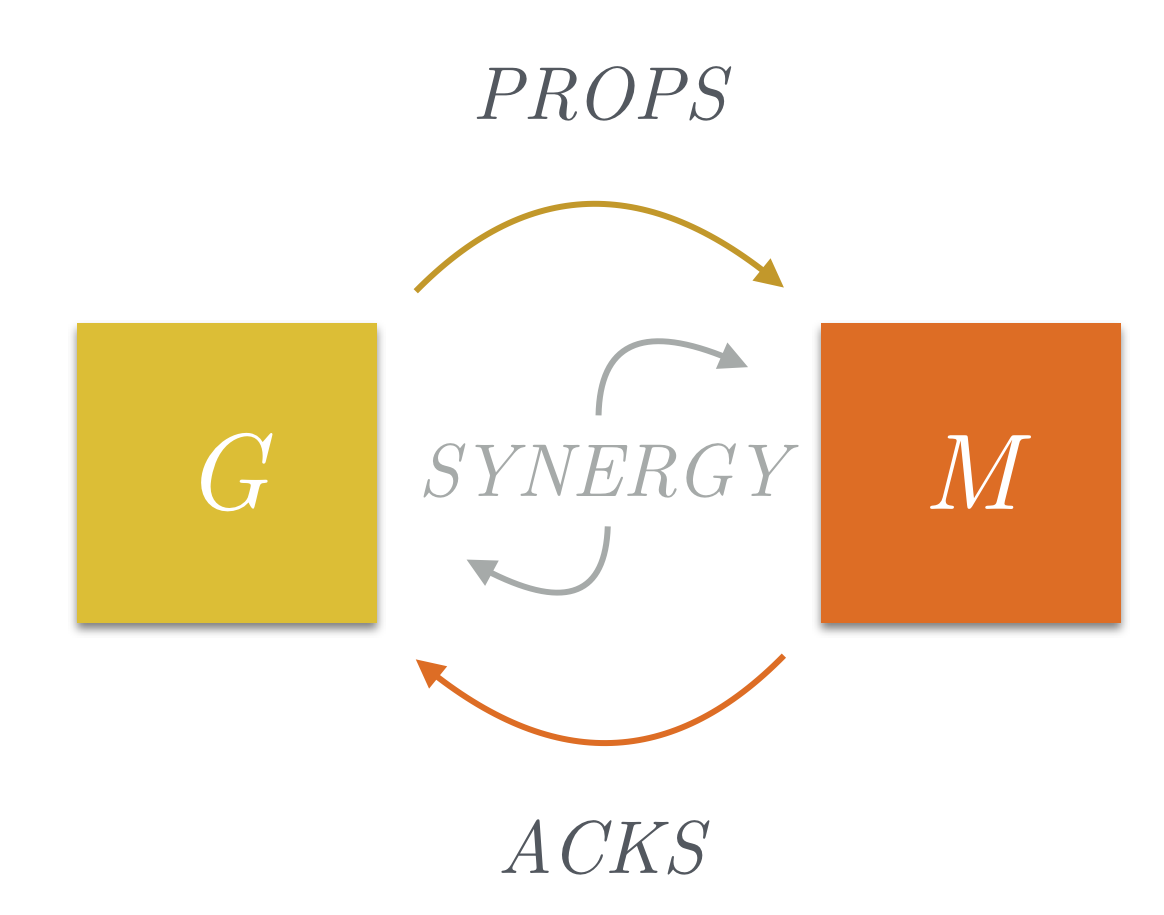}
    \caption{The proposed unadversarial training protocol. The generator $G$ 
        proposes samples, \textit{PROPS}, and in return receives 
        acknowledgements and praise, \textit{ACKS} from the motivator $M$. 
        As a direct consequence of the sense of teamwork fostered by our 
        optimisation scheme, \textit{synergy} abounds. Note: this figure 
        best viewed at a distance, preferably at low resolution. }
    \label{fig:training}
\end{figure}

Note that the equation above is valid and above board, since the current 
version of mathematics (v42.1 at the time of writing) supports simultaneous 
assignment\footnote{We caution readers not to rely on this assumption in 
future versions.  Mathematics has not supported backwards compatability since 
Kurt \textit{\say{Tab-Liebehaber}} G{\"o}del re-implemented the entire 
axiomatic foundations of the language rather than be constrained to four-space 
equation indentation (see \cite{godel1931formal} for the details).}. 
Therefore, in the following exposition, $D$ represents the generator and $G$ 
represents the discriminator. Next, we define a cost function, 
$C: \mathcal{V} \rightarrow \$$, mapping the space of model violence 
$\mathcal{V}$ into the space $\$$ spanned by all mattresses stuffed with 
U.S. dollars, as follows:

\begin{equation}
    C(V) = \alpha \int \beta_V(G)
\end{equation}

in which $\beta_V$ is a violent and discriminatory mapping from the 
discriminator $G$ to the closest mathematical structure which appears to be 
a human brain and $\alpha$ is a constant representing the cost of human 
violence, to be determined by trawling through posts on social media. 
Note that $\beta_V$ may be a violent function, but not crazy-violent (i.e. 
it must be \textit{Khinchin-integrable})\footnote{Since Neuroscience tells 
us that human brains are AlexVGGIncepResNets \textit{almost-everywhere}, in 
practice we found that these functions need not be overly belligerent.}.

To evaluate this cost, we first compute $\alpha$ with a melancholy search of 
Twitter, uniquely determining the cost of violence globally as $\$1876$ for 
every person in the world \citep{violenceTweet}. Integrating over all 
discriminators and cases of probable discrimination, we arrive at a 
conservative value of $3.2$ gigamattresses of cost. By any reasonable measure 
of humanity (financial, social, spiritual, cultural, grammatical or indeed 
dermatological), this is too many gigamattresses. 

Having made the compelling case for GUNs, we now turn to the highly anticipated 
 \textit{related work} section, in which we adopt a petty approach to resolving
 disagreements with other researchers by purposefully avoiding references to 
their relevant work.

\section{Related Work}

\epigraph{These violent delights have violent ends}{\textit{Geoff Hinton, 
date unknown}}

Our work is connected to a range of adversarial work in both the machine 
learning and the machine forgetting communities. To the best of our knowledge 
Smith \& Wesson (1852) were the first to apply GUNs to the problem of 
generative modelling, although similar ideas have been explored in the 
context of discriminative modelling as far back as the sixteenth century by 
Fabbrica d'Armi Pietro Beretta in an early demonstration of one-shot learning.  
Unfortunately, since neither work evaluated their approach on public 
benchmarks (not even on MNIST), the significance of their ideas remains 
under appreciated by the machine learning community.

Building on the approach of \cite{fouhey2012kardashian}\footnote{This 
innovative work was the first to introduce the concept of an 
alphabetically-related, rather than scientifically-related literature 
review.}, we next summarise the adversarial literature most closely related 
to ours, ordered by Levenshtein edit distance: GAN \citep{goodfellow2014generative}, 
WGAN \citep{arjovsky2017wasserstein}, DCGAN \citep{radford2015unsupervised}, 
LAPGAN \citep{denton2015deep}, InfoGAN \citep{chen2016infogan}, StackedGAN 
\citep{huang2016stacked} and UnrolledGAN \citep{metz2016unrolled}\footnote{In 
the interest of an unadversarial literature review, we note that 
\cite{bishop2006pattern} and \cite{murphy2012machine} make \textit{equally 
good} (up to $\epsilon=10^{-6}$) references for further exploration of this 
area.}. 

Unadversarial approaches to training have also received some attention, 
primarily for models used in other domains such as fashion 
\citep{crawford} and bodybuilding \citep{schwarzenegger2012arnold}).  
Some promising results have also been demonstrated in the generative modelling
domain, most notably through the use of \textit{Variational Generative 
Stochastic Networks with Collaborative Shaping} \citep{bachman2015variational}.  
Our work makes a fundamental contribution in this area by 
dramatically reducing the complexity of the paper title. 

\section{Generative Unadversarial Networks}

Under the Generative Unadversarial Network framework, we simultaneously train
two models: a generator $G$ that does its best to capture whichever data 
distribution it feels it can manage and a motivator $M$ that helps $G$ to 
achieve its dream. The generator is trained by learning a function 
$G(\vec{z};\theta_g)$ which transforms samples from a uniform prior 
 distribution $p_z(\vec{z})$ into the space graciously accommodating 
 the data\footnote{The choice of the uniform prior prevents discrimination 
 against prior samples that lie far from the mean. It's a small thing, 
 but it speaks volumes about our inclusive approach.}.  The motivator is 
 defined as a function $M(\vec{x};\theta_M)$ which uses gentle gradients
 and persuasive language to encourage $G$ to improve its game.  In particular,
 we train $G$ to maximise $log(M(G(\vec{z}))$ and we simultaneously train 
 $M$ to maximise $log(M(G(\vec{z}))$. Thus, we see that the objectives of 
 both parties are aligned, reducing conflict and promoting teamwork.

The core components of our framework are illustrated in Figure 
\ref{fig:training}. The GUN training scheme was inspired largely by Clint 
Eastwood's memorable performance in \textit{Dirty Harry} but also in part by 
the Transmission Control Protocol (TCP) three-way handshake 
\citep{postel1981transmission}, which was among the first protocols to build 
harmony through synergy, acknowledgements and the simple act of shaking hands.
A description of the training procedure used to train $G$ and $M$ is given in 
Algorithm \ref{trainingAlgo}. 
 
 \begin{figure}[t]
    \centering
    \includegraphics[width=.9\textwidth]{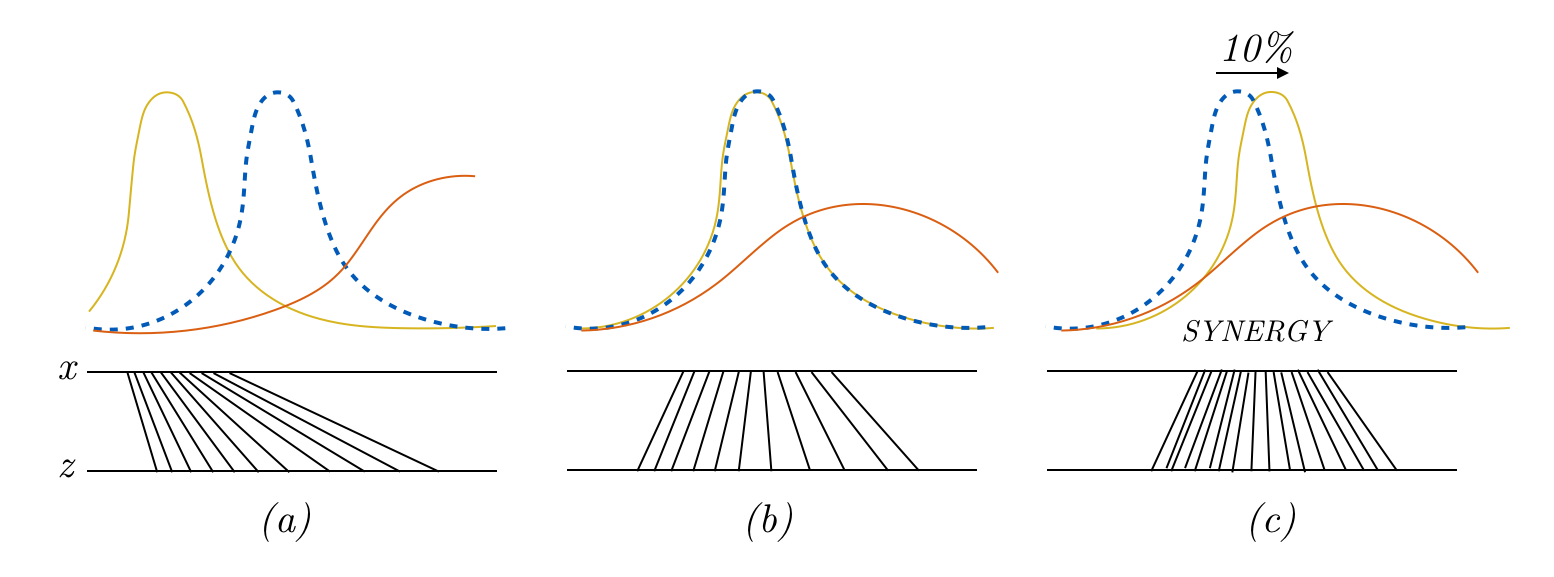}
    \caption{(a) GUNs are trained by updating the generator distribution $G$ 
    (yellow line) with the help and support of the motivator (red line) to 
    reach its dream of the data distribution (blue dashed). (b) With a 
    concerted effort, the generator reaches its goal. (c) Unlike previous 
    generators which were content with simply reaching this goal, our generator 
    is more motivated and gives it `$110\%$' moving it a further $10\%$
    past the data distribution. While this isn't terribly helpful from a 
    modelling perspective, we think it shows the right kind of attitude. 
    }
    \label{fig:mesh1}
\end{figure}

\begin{algorithm}[t]
  \caption{Training algorithm for Generative Unadversarial Networks}
  \label{trainingAlgo}
  \begin{algorithmic}[1]
    \Procedure{Train}{}
    \For{\texttt{\#iterations}}
        \State Sample $n$ noise samples from prior $p_z(\vec{z})$ and compute 
        $G({\vec{z}^{(1)};\theta_g), ...G({\vec{z}^{(n)};\theta_g)}}$.
        \State Sample $n$ data samples ${\vec{x}^{(1)}, ...\vec{x}^{(n)},}$ 
        from the data distribution.
        \State Let $G$ show pairs 
        $({\vec{x}^{(i)}}, G({\vec{z}^{(i)};\theta_g)})$ to $M$ as slides of a 
        powerpoint presentation\footnotemark.        
        \State Sample constructive criticism and motivational comments from $M$.
        \State Update the powerpoint slides and incorporate suggestions into 
        $\theta_G$.
      \EndFor
    \EndProcedure
  \end{algorithmic}
\end{algorithm} 
\footnotetext{To guarantee polynomial runtime, it is important to ensure that the 
generator is equipped with the appropriate dongle and works through any 
issues with the projector \textit{before} the presentation begins.}

Algorithm \ref{trainingAlgo} can be efficiently implemented by combining 
a spare meeting room (which must have a working projector) and a top notch 
deep learning framework such as MatConvNet \citep{vedaldi2015matconvnet} or 
Soumith Chintala (Chintala, 2012-present).   We note that we can further 
improve training efficiency by trivially rewriting our motivator objective 
as follows\footnote{If this result does not jump out at you immediately, 
read the odd numbered pages of \citep{amari2000methods}
. This book should be read in Japanese. The 
even-numbered pages can be ripped out to construct beautiful \emph{orizuru}.}:

\begin{equation}
    \theta_{M}^* = \underset{\theta_{M}}{\min} \oint\limits_{S(G)}\log(R) + \log(1-\zeta)
    \label{eqn:loss}
\end{equation}

Equation \ref{eqn:loss} describes the flow of reward and personal 
well-being on the generator network surface. $\zeta$ is a constant which 
improves the appearance of the equation.  In all our experiments, we fixed 
the value of $\zeta$ to zero.

\pagebreak

\section{Experiments}

\epigraph{Give the people what they want (MNIST)}{\textit{Yann LeCun, 
        date unknown}}

In this section we subject the GUN framework to a rigorous qualitative 
experimental evaluation by training unadversarial networks on MNIST. Rather 
than evaluating the model \textit{error-rate} or probability on 
\textit{withheld test data}, we adopt a less confrontational metric, 
\textit{opportunities for improvement}.  We also assess samples generated by 
the trained model by \textit{gut feeling}, enabling a direct comparison with 
a range of competing generative approaches. Following academic best practices, 
key implementation details can be found in our private code 
repository\footnote{We also make available a public copy of this repository 
which \textit{almost} compiles.  For the sake of brevity, all code comments, 
variables and function calls have been helpfully removed and replaced 
cross-platform, universally compatible ascii art. The code can be found at 
\url{http://github.com/albanie/SIGBOVIK17-GUNs}.}. 

We \textit{warm-start} the network with toy data taken from the 
latest Lego catalog. To nurture the right kind of learning environment, 
we let the network find its own learning rate and proceed by making 
$\epsilon$-greedy updates with an $\epsilon$ value of $1$. 
We consider hard-negative mining to be a gratuitously harsh training procedure,
and instead perform \textit{easy-positive mining} for gentler data digestion.

\begin{figure}
    \centering
    \includegraphics[width=0.45\textwidth]{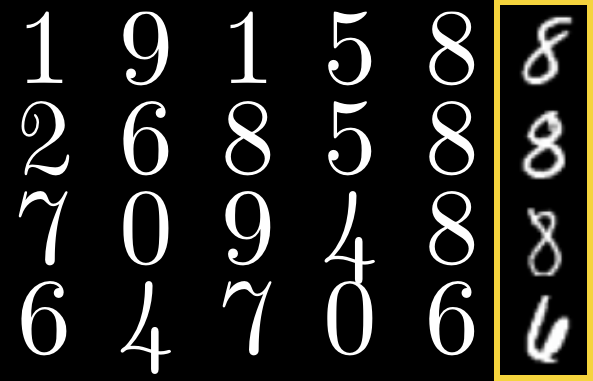}
    \caption[]{Visualised samples from the GUN model trained on 
    MNIST\protect\footnotemark (the nearest training examples are shown 
    in the right hand column).  Note that these samples have been carefully 
    cherry picked for their attractive appearance.  Note how the GUN samples 
    are much clearer and easier to read than the original MNIST digits.}
    \label{fig:mnist}
\end{figure}

We now turn to the results of the experiment.
Inspired by the Finnish education system, we do not test our models during the 
first formative epochs of development.
A quantitative comparison with two other popular 
generative approaches has been withheld from publication to respect the 
privacy of the models involved.  However, we are able to reveal that GUN had 
by far the most \textit{opportunities for improvement}.
We observed a sharp increase in performance once we all agreed that the 
network was doing well. By constrast, the adversarial nature of standard 
GAN methodologies usually elicits a fight-or-flight behavior, which can 
result in vanishing gradients and runaway losses.
Samples drawn from the trained network are shown in Figure \ref{fig:mnist}.

\footnotetext{For ease of visualisation, the GUN samples were lightly 
post-processed with \LaTeX.}

\section{Conclusion}

In this work, we have shown that network-on-network violence is not 
only unethical, it is also unnecessary.  Our experiments demonstrate 
that happy networks are productive networks, laying the groundwork for 
advances in motivational machine learning. Indeed, unadversarial 
learning is an area rife with opportunities for further development. 
In future work, we plan to give an expanded treatment of important related 
subjects including nurtural gradients and k-dearest neighbours\footnote{While 
we have exhaustively explored the topic of \textit{machine learning 
GUNs}, we leave the more controversial topic of \textit{machine GUN learning} 
to braver researchers.}.


\subsection*{Acknowledgements}

The authors would like to acknowledge the quality of Karel Lenc's homemade pancakes. This work was supported by the NRA (National Research Association). 



\bibliography{iclr2016_conference}
\bibliographystyle{iclr2016_conference}

\section*{Authors' Biographies}
\subsection*{Samuel}
Samuel started writing biographies at the tender age of 24, when he penned 
his first short story \say{Ouch that seriously hurt, keep your **** cat away 
from me} about the life of Jack Johnson, his brother's lovable albino cat 
with anger management issues. His career as a biographer has gone from 
strength to strength ever since, flourishing in several other phyla of the 
animal kingdom. He is a noted expert on the much beloved native English 
Panda and is a self-award winning author on the challenges of hunting them. 

\subsection*{Sebastien}
\sethlcolor{black}
Sebastien holds a self-taught liberal arts degree, and passed his driver's 
license exam with highest honours. Secretly a \hl{German} national, he then 
joined the French Foreign Legion and was deployed \hl{Sam-- Don't think 
that the redacted joke is funny enough to justify the loss of a biography - 
will return to this later. Interestingly, this latex package does not 
redact full stops.  Possible gap in the market here?} in Nicaragua, \hl{~I~} 
of \hl{like}, \hl{turtles}.\\
\hl{Joao-- Fixed it. Actually I quite like this bio, it has a certain 
mysterious quality to it. I wonder if there's a way to hack the PDF and 
read what's written underneath. In any case I'm just gonna write my 
groceries list here so I can easily access it on my phone when I'm in the 
shop, hope you don't mind.\\
- Eggs\\
- Milk\\
- Ammo\\
}

\subsection*{Jo\~ao}
Jo\~ao \textit{El Tracko} F. Henriques holds a joint bachelors degree in 
guerilla warfare tactics and cakemaking from the University of Coimbra, 
where he has been tracking down the \st{victims} subjects of his critically 
acclaimed biographies for over five years. Little did they know that his visual
object tracking skills extend to real-life. Though some (all) of his 
subjects have since passed away, their legend lives on his thoughtfully
written monograph, \say{How to most effectively interview someone who is 
trying desperately to escape from you}.

\end{document}